\documentclass[letterpaper, 10 pt, conference]{ieeeconf}  % Comment this line out if you need a4paper

\IEEEoverridecommandlockouts                              % This command is only needed if 
\usepackage{graphicx} % for pdf, bitmapped graphics files
\usepackage{amsmath} % assumes amsmath package installed
\usepackage{amssymb}  % assumes amsmath package installed
\usepackage{xfrac}

\title{\LARGE \bf
Triple Scissor Extender: A 6-DOF Lifting and Positioning Robot*
}

\author{Daniel J. Gonzalez and H. Harry Asada$^{1}$% <-this % stops a space
\thanks{*This work was supported by The Boeing Company.}% <-this % stops a space
\thanks{$^{1}$The authors are with the d'Arbeloff Laboratory for Information Systems and Technology in the Department of Mechanical Engineering,
        Massachusetts Institute of Technology, Cambridge, MA 02139, USA. Email: 
        {\tt\small \{dgonz, asada\}@mit.edu}}%
}

\begin{document}

\maketitle
\thispagestyle{empty}
\pagestyle{empty}

%%%%%%%%%%%%%%%%%%%%%%%%%%%%%%%%%%%%%%%%%%%%%%%%%%%%%%%%%%%%%%%%%%%%%%%%%%%%%%%%
\begin{abstract}
We present a novel 6 DOF robotic mechanism for reaching high ceilings and positioning an end-effector. The end-effector is supported with three scissor mechanisms that extend towards the ceiling with 6 independent linear actuators moving the base ends of the individual scissors. The top point of each scissor is connected to one of three ball joints located at the three vertices of the top triangular plate holding the end-effector. Coordinated motion of the 6 linear actuators at the base allows the end-effector to reach an arbitrary position with an arbitrary orientation. The design concept of the Triple Scissor Extender is presented, followed by kinematic modeling and analysis of the the Inverse Jacobian relating actuator velocities to the end-effector velocities. The Inverse Jacobian eigenvalues are determined for diverse configurations in order to characterize the kinematic properties. A proof-of-concept prototype has been designed and built. The Inverse Jacobian for use in differential control is evaluated through experiments.


\end{abstract}

%%%%%%%%%%%%%%%%%%%%%%%%%%%%%%%%%%%%%%%%%%%%%%%%%%%%%%%%%%%%%%%%%%%%%%%%%%%%%%%%
\section{Introduction}
Industrial automation applications requiring both a high payload capacity and a large workspace typically rely on large serial link articulated robots. While these robots are an excellent choice for factory floors with ample maneuvering space, they are often too heavy for mobile applications and unable to reach the desired workspace in confined settings. 

For example, a stationary articulated robot arm can easily perform various operations along the outside of a commercial aircraft fuselage as it is being assembled, but this same arm cannot be placed on a mobile base and rolled \textit{into} the fuselage barrel to perform additional operations: the arm is simply too large and heavy to make this usage feasible. 

Parallel manipulators such as the 6-DOF Gough-Stewart hexapod platform \cite{RGough} \cite{Rstewart}, are small and light relative to their load-bearing capacity, unlike articulated serial-link arms. These advantages make the 6-DOF parallel manipulator a candidate for confined-space maneuvering and assembly, but they have small workspaces due to the use of piston-style prismatic actuators as their linkages. A fully-retracted Parallel Platform of unit height $h$ cannot reach past $2h$ because its actuators cannot extend any more than twice their smallest length.

   \begin{figure}[ht] %thpb
   	\centering
   	\includegraphics[scale=0.2]{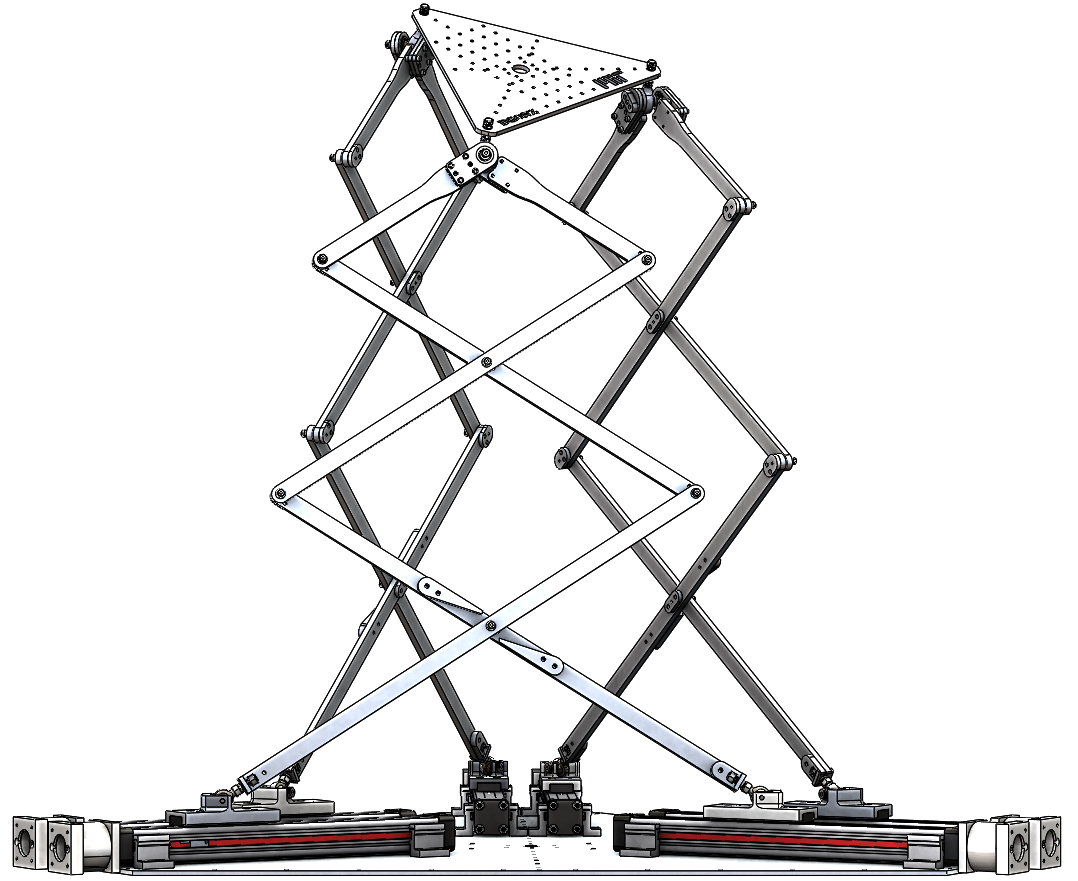}
   	\caption{Triple Scissor Extender}
   	\label{figurelabel}
   \end{figure}   

To address this limitation, scissor mechanisms can be used to amplify the height range. A 3-DOF example of this concept is analyzed in \cite{RYangWang}. We designed a new class of 6-DOF parallel platform that uses parallel scissor mechanisms to achieve a large workspace compared to its original size while maintaining the benefits of most parallel manipulators. Three scissor lift mechanisms are combined so that the end-effector can be supported by them in parallel and the position and orientation of the end-effector can be controlled arbitrarily in 3-dimensional space.

\begin{figure}[ht]
	\centering
	\includegraphics[scale=.2]{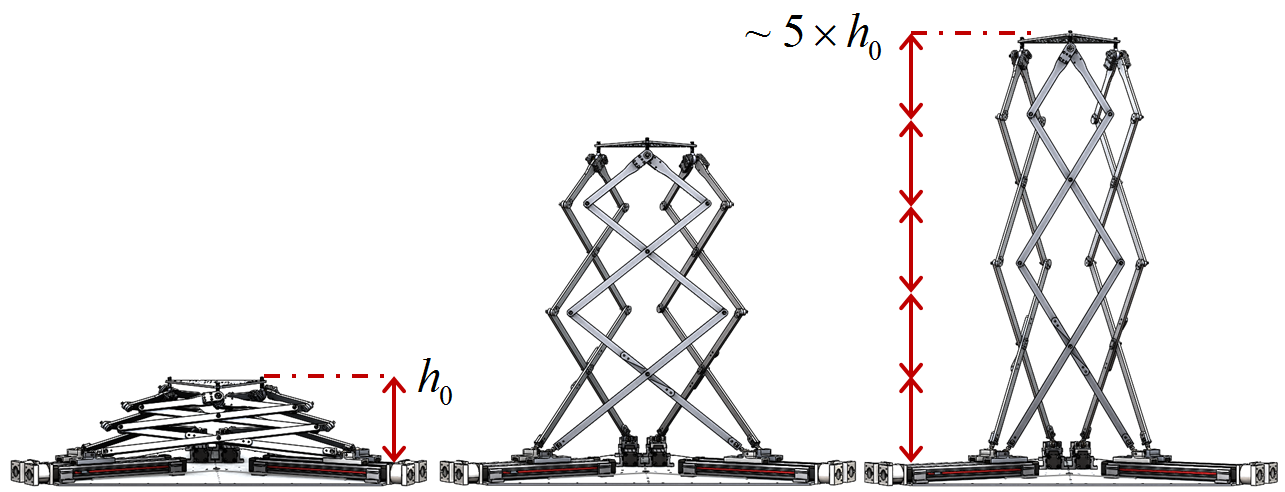}
	\caption{Demonstration of the Triple Scissor Extender's large height range}
	\label{figurelabel}
\end{figure}

This Triple Scissor Extender (TSE) (see Fig. 1) is particularly useful both for reaching high ceilings or walls and positioning/orienting its end-effector freely within a working space. When the TSE is fully contracted, the whole structure becomes compact, making it easy for transportation. At full extension, the device height becomes many times greater than its original height (like a traditional 1-DOF aerial scissor lift commonly used for maintenance, assembly, and construction), allowing it to reach high ceilings and walls. 

We first explain in detail the concept of extending the workspace of a 6-DOF parallel manipulator with pantograph/scissor mechanisms and the design of the TSE. The relationship between the 6 actuator inputs and the 6-DOF 3D pose, that is, the combined position and orientation, of the top platform are revealed through the Kinematic Constraint Equations.

The Inverse Jacobian Matrix, which characterizes the linear differential relationship between the inputs and outputs of the TSE system about a home position, is derived from the Kinematic Constraint Equations. A Singular Value Decomposition is performed on the Inverse Jacobian Matrix in order to reveal important properties of the TSE about these home positions. 

We then reveal our prototype of the TSE, perform an experimental validation of positioning accuracy using our Inverse Jacobian control scheme, and discuss the results.

\section{Design Concept}
This design is a combination of two concepts: the use of the pantograph or scissor mechanism to amplify motion and the kinematics of 6-DOF parallel manipulators like the Stewart-Gough platform. 
%
%\begin{figure}[ht]
%	\centering
%	\includegraphics[scale=0.15]{diagrams/Fig1}
%	\caption{Traditional Scissor Lift. }
%	\label{figurelabel}
%\end{figure}   

Consider two links of length $\ell_0$ existing on a plane $xy$ joined together at one end with a rotational joint at point $C$. The other end of each link is attached to a rotational joint that is coupled to a linear slide located on the $x$-axis. Both of these linear slides $s_A$ and $s_B$ travel along the same line. The point we wish to control is point C, which has 2 degrees of freedom, $(x_C,y_C)$. As inputs, we can change the position, $x_A$ or $x_B$ of each linear slide. Two modes of motion exist: if both slide $s_A$ and slide $s_B$ move at the same velocity along the $x$-axis, then point $C$ will also move at that same velocity in the $x$ direction; if slide $s_A$ and slide $s_B$ move towards or away from each other with equal and opposite velocities, then point $C$ will move only in the $y$ direction. Through the superposition of these modes of motion, we can reach any point above the $x$-axis up to some maximum height.

Now consider extending each link beyond point $C$ by some smaller length $\ell_1$, each with a rotational joint at the end, and then adding to those joints two more links of length $\ell_1$ that are joined together at the other end, which becomes the new point $C$. We now have a triangle with a parallelogram on top: a basic scissor mechanism. The same two modes of motion exist as in the previous case but the ratio of inward motion of slides $s_A$ and $s_B$ and the vertical motion of point C has been amplified by the pantograph mechanism! Additional parallelograms of side length $\ell_2\textless \ell_3\textless \dots\textless \ell_n$ can be added to the assembly to produce a mechanism like that in Fig. 3.

   \begin{figure}[ht]
   	\centering
   	\includegraphics[scale=0.25]{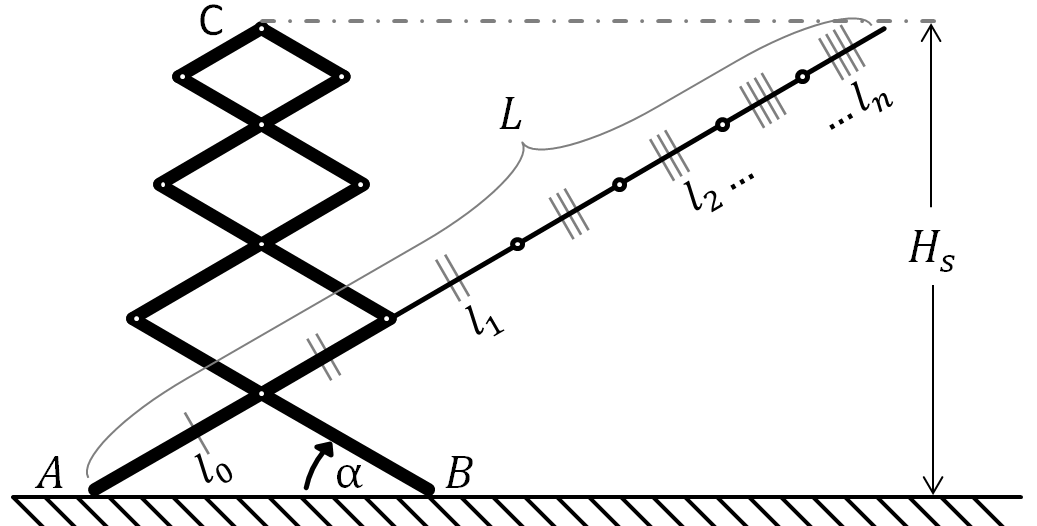}
   	\caption{Kinematic parameters of a single scissor mechanism. }
   	\label{figurelabel}
   \end{figure} 

Now we introduce the third dimension to the model and change the coordinate frame orientation such that the previous $xy$-plane becomes the new $yz$-plane, with the $z$-axis pointed upwards, the points $A$ and $B$ lying along the $y$-axis, and the $x$-axis orthogonal to the $yz$-plane. We turn the rotational joints at slide $s_A$ and slide $s_B$ into ball joints that allow the entire scissor mechanism to rotate about the line $\overline{AB}$. Ball joints are used because the scissor mechanism is required to pitch and yaw with respect to the linear slides in 3D space while moving, and must roll with respect to the linear slides in order to rotate about line $\overline{AB}$. 

Now consider the scissor mechanism's location on the new $xy$-plane. We introduce two identical scissor mechanisms, which we label $2$ and $3$, and then arrange the trio in a triangular configuration. We introduce a small triangular top plate, and connect points $C_1$, $C_2$, and $C_3$ to the three apices of the top plate via ball joints. Finally we can put the six independent linear slides $s_{A1}$, $s_{B1}$, $s_{A2}$, $s_{B2}$, $s_{A3}$, and $s_{B3}$ (two per scissor) on a plane and arrange them in pairs to get the final formulation of the TSE, shown in Fig. 1.

\begin{figure}[ht]
	\centering
	\includegraphics[scale=.32]{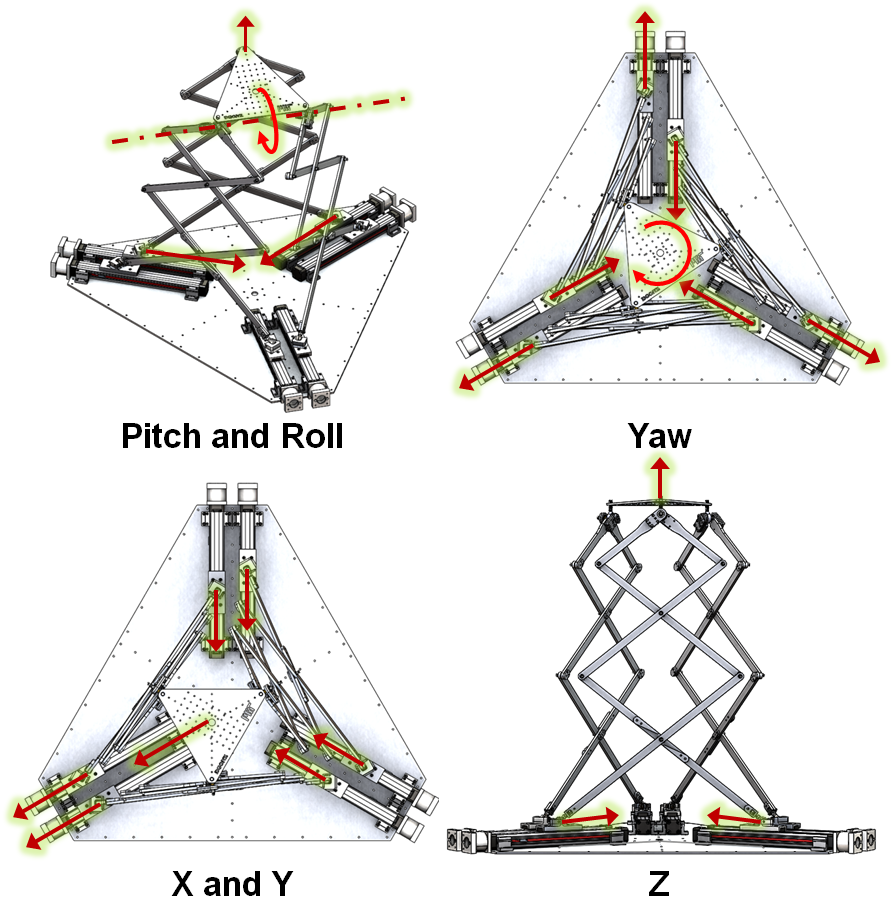}
	\caption{Intuitive Coordinated Motion Patterns of the Triple Scissor Extender}
	\label{figurelabel}
\end{figure} 

The top platform of the TSE has 6 DOFs. Some of these can be deduced intuitively by reasoning about the differential motion of the six slides and how they affect the pose of the top platform (see Fig. 4). By moving only slides $s_{A1}$ and $s_{B1}$ inwards, point $C_1$ moves upwards, rotating the top platform about line $\overline{C_2C_3}$. 
Similar rotations can be achieved with scissor mechanisms $2$ and $3$. Through a combination of these motions, the pitch and the roll of the top platform can be commanded. By moving all 6 slides inward simultaneously, the top platform translates upward. By alternating the direction of each slide sequentially, the top plate rotates  about the $z$-axis in a yaw motion. By moving one adjacent pair of scissors outward and the remaining scissors inward, translational motion is achieved in the average direction of motion. 

By the coordinated superposition of these various modes of motion, the top platform can move to any desired pose within its workspace. 

\section{Kinematic Modeling}
Traditional 6-DOF platforms have relatively simple inverse kinematics solutions \cite{RKine}, but the highly coupled motion of the scissors relative to the actuators in the TSE make finding the inverse kinematics challenging. We now analyze the kinematic behavior of the TSE subject to geometric constraints and attain kinematic constraint equations. 

Fig. 5 shows the coordinate system used for describing the kinematic behavior of the Triple Scissor Extender. The top plate position is represented with vector $X_{\epsilon}^T=\begin{pmatrix} x_{\epsilon}& y_{\epsilon}& z_{\epsilon}\end{pmatrix}^T$ with reference to the base coordinate system $O-xyz$. The orientation of the top plate is described with roll, pitch, and yaw angles $\Theta=\begin{pmatrix}\varphi & \theta & \psi \end{pmatrix}^T$. 

\begin{figure}[ht]
	\centering
	\includegraphics[scale=0.32]{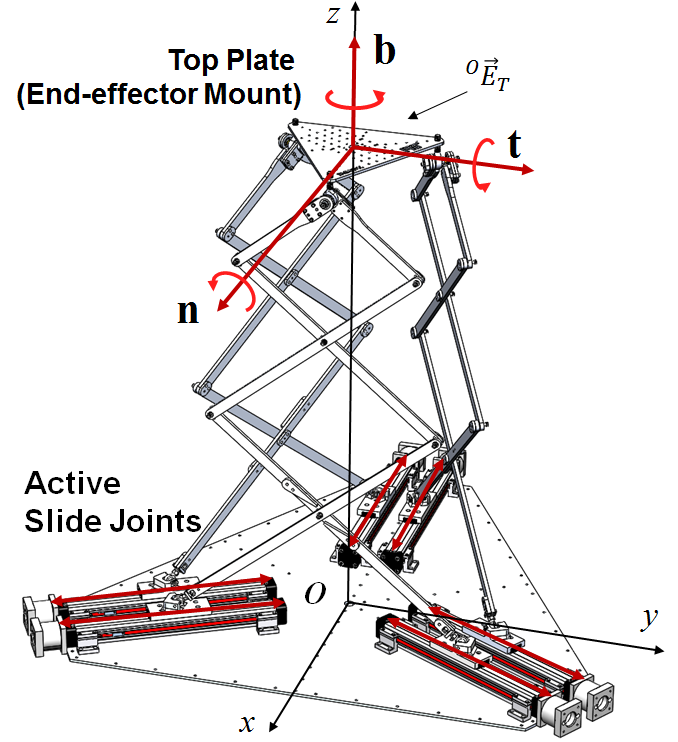}
	\caption{Key variables and parameters of the base platform. }
	\label{figurelabel}
\end{figure}   

As shown in Fig. 5, let $\hat{n}$, $\hat{t}$, $\hat{b}$ be, respectively, the unit vectors pointing in the three directions of a Cartesian coordinate frame $O_{\epsilon}-x'y'z'$ attached to the top plate. Concatenating these unit vectors in a 3x3 matrix we can write the orientation of the top plate in a compact expression
\begin{equation}
\begin{bmatrix} \hat{n} & \hat{t} & \hat{b} \end{bmatrix}
=R_{xyz}(\varphi,\theta,\psi)=
R_x(\varphi)
R_y(\theta)
R_z(\psi)
\end{equation}
where $R_x(\varphi)$, $R_y(\theta)$, $R_z(\psi)$ are 3x3 rotation matrices about the x, y, and z axes, respectively.

The three apices of the top plate, $C_1$, $C_2$, and $C_3$, are connected to the three independent scissor mechanisms, while each scissor mechanism is activated with two linear actuators at the base. Let $s_{Ai}$ and $s_{Bi}$ be displacements of the linear actuators moving points $A_i$ and $B_i$ of the $i$-th scissor mechanism. Collectively, the six actuator displacements:
\begin{equation}
q=\begin{pmatrix}s_{A1} & s_{B1} & s_{A2} & s_{B2} & s_{A3} & s_{B3}\end{pmatrix}^T
\end{equation}
form a 6-dimensional joint coordinate vector. These joint coordinates determine the top plate position and orientation:
\begin{equation}
p=\begin{pmatrix}X_{\epsilon}^T & \Theta_{\epsilon}^T\end{pmatrix}^T=
\begin{pmatrix}x_{\epsilon}& y_{\epsilon}& z_{\epsilon}& \varphi_{\epsilon} & \theta_{\epsilon} & \psi_{\epsilon}\end{pmatrix}^T
\end{equation}

The kinematic equation relating the endpoint pose $p$ to the actuator displacements $q$ is prohibitively complex, while its inverse kinematic relationship is tractable: $q = f (p)$. 

   \begin{figure}[ht]
   	\centering
   	\includegraphics[scale=0.3]{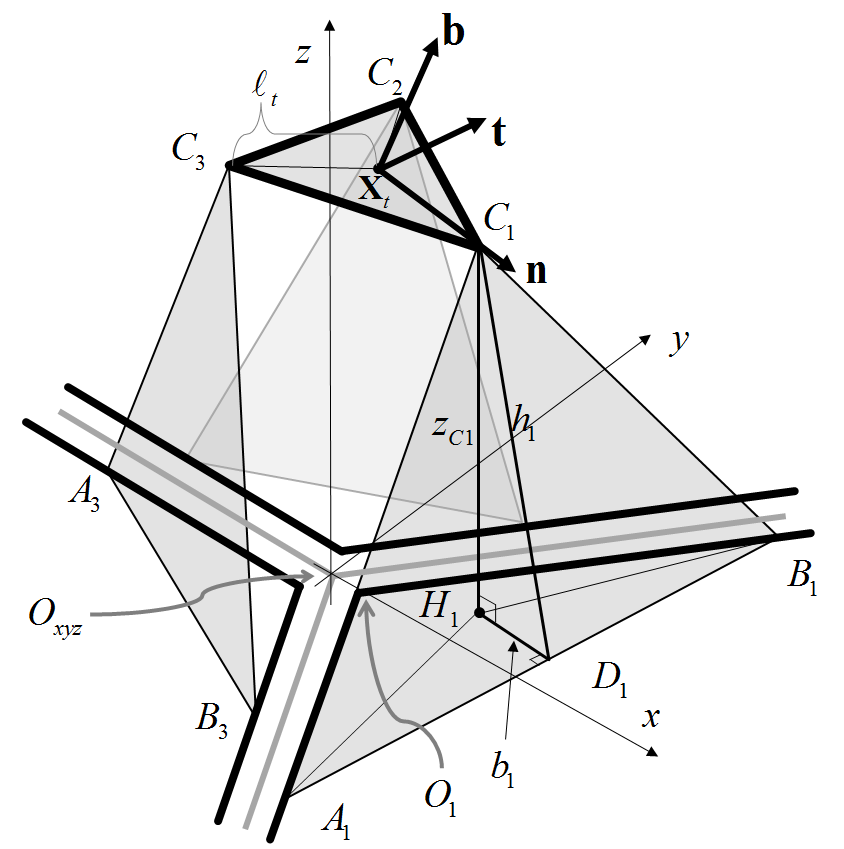}
   	\caption{Projection of Relevant Points onto the Base Plane}
   	\label{figurelabel}
   \end{figure}
   
The inverse kinematics problem can be solved in the following steps:

\begin{itemize}
	\item
	Given the 6-DOF pose $p$ of the top plate, obtain coordinates of points $C_1$, $C_2$, and $C_3$ relative to the $O-xyz$ frame using the unit vectors in (1)
	\begin{equation}
	\begin{split}
	X_{C1}&=X_t+\ell_t\hat{n}\\
	X_{C2}&=X_t-\ell_t\left(\frac{\sqrt{3}}{2}\hat{t}+\frac{1}{2}\hat{n}\right)\\ X_{C3}&=X_t+\ell_t\left(\frac{\sqrt{3}}{2}\hat{t}-\frac{1}{2}\hat{n}\right)
	\end{split} %Use ampersand "&" in order to align the split!
	\end{equation}
	where $\ell_t$ is the distance from the center of the top plate to each apex (See Fig. 6).
	\item For each scissor mechanism $i$, given top point coordinates $C_i$, solve for the actuator displacements $(s_{Ai}, s_{Bi})$. In order to maintain generality, an intermediate coordinate system $O_i-xyz$ is created with its origin located at the intersection of linear slides $A$ and $B$ (See Fig. 6), its $O_i-z$ axis parallel to the $O-z$ axis, and its $y$-axis pointing outward from the center of the TSE (see Fig. 7). The point coordinates of $C_i$ are converted from the TSE origin frame $O-xyz$ to the new frame $O_i-xyz$, and the general Kinematic Constraint Equations derived in the next sections can be used to obtain the actuator displacements $(s_{Ai}, s_{Bi})$
\end{itemize}

We obtain the Kinematic Constraint Equations in the following subsections.

\subsection{2D Scissor Mechanism}
First, the basic 2D kinematic relationship of a single scissor mechanism will be obtained. As shown in Fig. 3, there is a functional relationship between the width of the scissors base, $w_i=\overline{A_iB_i}$, and the height of the scissors $h_i$, $i=1,2,3$. For brevity, the subscript $i$ will be omitted hereafter. The scissor mechanism consists of $n$ parallelograms of side length $\ell_1, \ell_2,\dots,\ell_n$, and one isosceles triangle of equal side $\ell_0$, connected at the center nodes $N_1, N_2,\dots, N_n$. Let $\alpha$ be the angle of each scissor link relative to the baseline, $\alpha=\angle ABN_1$, as shown in the figure. The width $w$ is given by
\begin{equation}
w=2\ell_0cos(\alpha)
\end{equation}
Since all the links are kept parallel, the height h is given by
\begin{equation}
h=Lsin(\alpha)
\end{equation}
where the total length L is given by
\begin{equation}
L=\ell_0+2\ell_1+\dots+2\ell_{n-1}+2\ell_n
\end{equation}
See Fig. 3 for geometric interpretation. Eliminating angle $\alpha$ from (5) and (6) yields
\begin{equation}
\left(\frac{h}{L}\right)^2+\left(\frac{w}{2\ell_0}\right)^2=1
\end{equation}

\subsection{Projection onto the $O_1-xy$ Plane}
As shown in Fig. 6, consider the projection of Point $C_1$ onto the base plane, $O_1-xy$. The projected point $H_1$ is redrawn within the $O_1-xy$-plane in Fig. 7. Again, for brevity, the subscript $1$ is omitted in the following equations.

The $(x,y)$ coordinates of scissor base points $A$ and $B$ are determined by the linear actuators, which move the scissors base points along the two radial directions, respectively. Therefore,
\begin{equation}
	\begin{split}
	x_A=s_Acos\left(\frac{\pi}{6}\right),\ \ &
	y_A=s_Asin\left(\frac{\pi}{6}\right)\\
	x_B=-s_Bcos\left(\frac{\pi}{6}\right),\ \ &
	y_B=s_Bsin\left(\frac{\pi}{6}\right)
	\end{split}
\end{equation}
\begin{figure}[ht]
	\centering
	\includegraphics[scale=0.28]{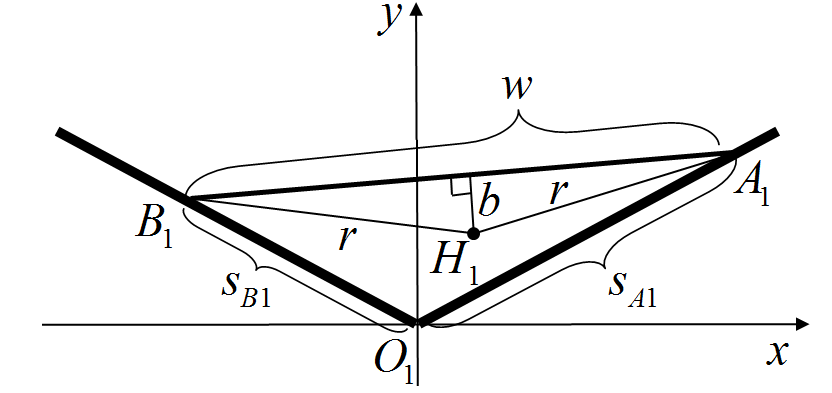}
	\caption{Top View of Base Plane showing Projected Point $H$ on $O_1xy$-plane}
	\label{figurelabel}
\end{figure}
Using these coordinates, the width of the scissors, $w$, can be written as
\begin{equation}
	w^2=(x_A-x_B)^2+(y_A-y_B)^2
\end{equation}
which, when combined with (9) simplifies to 
\begin{equation}
	w^2=s_A^2+s_Bs_A+s_B^2
\end{equation}
Note that the scissor mechanism is symmetric with respect to its centerline connecting the center nodes $N_1, N_2,\dots, N_n$. Therefore, point $H$, that is, the projection of point $C$ onto the $xy$-plane, is on the bisector of the baseline $AB$. Hence, $\overline{AH}=\overline{BH}=r$, or
\begin{equation}
	\begin{split}
	r^2&=(x_A-x_C)^2+(y_A-y_C)^2\\
	&=(x_B-x_C)^2+(y_B-y_C)^2
	\end{split}
\end{equation}
where $x_C$ and $y_C$ are the $xy$ coordinates of point $C$, i.e. those of point $H$. This produces two conditions:
\begin{equation}
	s_A^2-\sqrt{3}(s_A+s_B)x_C=s_B^2+(s_A-s_B)y_C
\end{equation}
\begin{equation}
	r^2=s_A^2-s_A(\sqrt{3}x_C+y_C)+x_C^2+y_C^2
\end{equation}

\subsection{Implicit equations relating apex coordinates $(x_C, y_C, z_C)$  and actuator displacements $(s_A, s_B)$ } 
   
The $z$ coordinate of point $C$ provides another condition. Considering the right triangle $C_1H_1D_1$ in Fig. 6, we obtain
\begin{equation}
	h^2=b^2+z_C^2
\end{equation}
From Fig. 7,
\begin{equation}
	r^2=b^2+\left(\frac{w}{2}\right)^2
\end{equation}
Eliminating $r$, $w$, $h$, and $b$ from (11), (14), (15) and (16) yields
\begin{equation}
	\begin{split}
	x_C^2+y_C^2+z_C^2=L^2-s_A^2+s_A(\sqrt{3}x_C+y_C)+\\
	\frac{1}{4}\left(1-\left(\frac{L}{\ell_0}\right)^2\right)(s_A^2+s_Bs_A+s_B^2)
	\end{split}
\end{equation}
Simultaneous equations (13) and (17) determine actuator displacements $(s_A, s_B)$ for given coordinates $(x_C, y_C, z_C)$. While the above derivation was only carried out for the first scissor mechanism, the other two scissor mechanisms can be treated in the same manner. 

\section{The Inverse Jacobian and Singular Value Analysis}

\subsection{Characterizing the TSE Kinematic Properties} 

Based on the Inverse Kinematic relations obtained in the previous section, basic properties of the Triple Scissor Extender will be highlighted in this section through the analysis of differential motion. 

Given a limited movable range, or stroke, of each actuator, ``small workspace'' implies a small end-effector displacement relative to large displacements at the actuators. In other words, the ratio of the end-effector displacement to the actuator displacements is small. Let $\left| \Delta q\right|=\sqrt{\Delta s_{A1}^2+\Delta s_{A1}^2+\dots+\Delta s_{B3}^2}$  be the magnitude of the six actuator displacements, and $\left|\Delta X_\epsilon\right|$ and $\left|\Delta \Theta_\epsilon\right|$ be, respectively, the magnitude of the translational and rotational displacements of the end-effector. We characterize the kinematic properties of the TSE in terms of the ratios:
\begin{equation}
\gamma_t(p)=\frac{\left|\Delta X_\epsilon\right|}{\Delta q}, \gamma_r(p)=\frac{\left|\Delta \Theta_\epsilon\right|}{\Delta q}, 
\end{equation}
which physically mean spatial, multi-DOF ``gear ratios" associated with the translation and rotation of the end-effector, respectively. These ratios vary depending on the end-effector pose $p$, as well as the direction of the end-effector motion. We will obtain the maximum and the minimum of $\gamma_t(p)$ and $\gamma_r(p)$ at each configuration of the end-effector pose $p$, and examine how the maximum/minimum ratios distribute over the workspace. 

This entails identifying the Jacobian relating the end-effector displacements to actuator displacements 
\begin{equation}
\Delta q = \mathbb{J}_{I} \Delta p
\end{equation}
where $\mathbb{J}_I$ is the $6\times6$ Inverse Jacobian matrix. This is also often called the Manipulability Matrix \cite{RManipulability}. Note that the elements of the Jacobian Matrix are partial derivatives of input motion to output motion, not time derivatives.

As in the previous kinematic analysis, the Inverse Jacobian Matrix can be split into two main parts. We obtain the differential relationship 
\begin{equation}(\Delta C_1 \quad \Delta C_2 \quad \Delta C_3)^T=\mathbb{J}_C\Delta\vec{p}\end{equation}
where $\mathbb{J}_C$ is the $9\times6$ Jacobian between the top platform pose $\vec{p}$ and the top platform apexes $C_i$. For top platform translation, differential motion is 1:1, and the left half of  $\mathbb{J}_C$ is made of Identity matrices. For top platform rotation, which corresponds with the right half of $\mathbb{J}_C$, the differential motion can be obtained from the transform matrices of (1). 

For each scissor we obtain the differential relationship between point $C_i$ and actuator displacements $s_A$ and $s_B$ where
\begin{equation}
\mathbb{J}_{S_i} = 
\begin{bmatrix} 
\cfrac{\partial s_{Ai}} {\partial x_{Ci}}  &
\cfrac{\partial s_{Ai}} {\partial y_{Ci}} &
\cfrac{\partial s_{Ai}} {\partial z_{Ci}}
\\ 
\cfrac{\partial s_{Bi}} {\partial x_{Ci}}  &
\cfrac{\partial s_{Bi}} {\partial y_{Ci}} &
\cfrac{\partial s_{Bi}} {\partial z_{Ci}}
\end{bmatrix}
\end{equation}
is the individual scissor Jacobian and the block matrix
\begin{equation}
\mathbb{J}_{S} = 
\begin{bmatrix}\mathbb{J}_{ S1 }\quad 0 \quad 0\\ 0 \quad \mathbb{J}_{ S2 } \quad 0 \\ 0 \quad 0 \quad \mathbb{J}_{ S3 }\end{bmatrix}
\end{equation}
is the combined $6\times9$ scissor Jacobian.

When both $\mathbb{J}_C$ and $\mathbb{J}_S$ combined, we obtain
\begin{equation}
\mathbb{J}_{I} = \mathbb{J}_{ S } \mathbb{J}_C
\end{equation}
which is the final $6\times6$ Inverse Jacobian Matrix. 

%(x_C1, y_C1, z_C1,x_C2, y_C2, z_C2,x_C3, y_C3, z_C3)
Obtaining $\mathbb{J}_{S}$ typically requires the Inverse Kinematics, which is not explicitly solvable. In the following sections the outline of an alternative computation of $\mathbb{J}_S$ is described.

\subsection{Computation of the Jacobians}  

TSE consists of three pairs of scissor mechanisms, each governed by the implicit kinematic equations (13) and (17). These determine the relationship between actuator displacements $s_A$, $s_B$ and the apex position $x_C$,  $y_C$,  $z_C$ in the local coordinate frame $O_i - x_iy_iz_i$ . For brevity the subscript is again omitted. Differentiating both equations (13) and (17) at a given apex position $p$, we can obtain a differential relationship in the following form:

\begin{equation}
\begin{split}
& {{a}_{11}}\Delta {{s}_{A}}+{{a}_{12}}\Delta {{s}_{B}}={{b}_{11}}\Delta {{x}_{c}}+{{b}_{12}}\Delta {{y}_{c}}+{{b}_{13}}\Delta {{z}_{c}} \\ 
& {{a}_{21}}\Delta {{s}_{A}}+{{a}_{22}}\Delta {{s}_{B}}={{b}_{21}}\Delta {{x}_{c}}+{{b}_{22}}\Delta {{y}_{c}}+{{b}_{23}}\Delta {{z}_{c}} \\ 
\end{split}
\end{equation}
where parameters $a_{11}, \dots, b_{23}$ are evaluated at the given apex position $\vec{p}$. Using vector-matrix form, we can write the Jacobian relating the two actuator displacement $\Delta s_A$, $\Delta s_B$ to those of the apex coordinates  $\Delta x_C$,  $\Delta y_C$,  $\Delta z_C$ as:
\begin{equation}
\left( \begin{matrix}
\Delta {{s}_{A}}  \\
\Delta {{s}_{B}}  \\
\end{matrix} \right)={{\mathbb{J}}_{S}}\Delta {{\mathbf{X}}_{c}},\quad \Delta {{\mathbf{X}}_{c}}=\left( \begin{matrix}
\Delta {{x}_{c}}  \\
\Delta {{y}_{c}}  \\
\Delta {{z}_{c}}  \\
\end{matrix} \right)
\end{equation}

where  ${{\mathbb{J}}_{S}}={{\mathbf{A}}^{-1}}\mathbf{B},\ \mathbf{A}=\left\{ {{a}_{ij}} \right\},\mathbf{B}=\left\{ {{b}_{ij}} \right\}$. Of particular interest is the case where the top plate is kept level and moves along the $z$-axis. At this center configuration, ${{s}_{A}}={{s}_{B}}=s$ and ${{x}_{c}}=0,\ {{y}_{c}}={{\ell }_{t}}-\Delta \gamma $ (where $\gamma$ is half the distance between adjacent linear slides such as $s_{B1}$ and $s_{A2}$), the Jacobian is given by
\begin{equation}
\begin{split}
&{{\mathbb{J}}_{S}}\left| _{Center} \right.=\\
&\left( \begin{matrix}
\dfrac{\sqrt{3}s}{2s-{{\ell }_{t}}+\Delta \gamma } & -\dfrac{\sqrt{3}s}{2s-{{\ell }_{t}}+\Delta \gamma } \\ \dfrac{2({{\ell }_{t}}-\Delta \gamma )-s}{6cs-2s+{{\ell }_{t}}-\Delta \gamma } & \dfrac{2({{\ell }_{t}}-\Delta \gamma )-s}{6cs-2s+{{\ell }_{t}}-\Delta \gamma } \\ \dfrac{2{{z}_{c}}}{6cs-2s+{{\ell }_{t}}-\Delta \gamma }
& \dfrac{2{{z}_{c}}}{6cs-2s+{{\ell }_{t}}-\Delta \gamma }  \\
\end{matrix} \right)^T
\end{split}
%\begin{split}
%&{{\mathbb{J}}_{C}}\left| _{Center} \right.=\\
%&\left( \begin{matrix}
%\dfrac{\sqrt{3}s}{2s-{{\ell }_{t}}+\Delta \gamma } & \dfrac{2({{\ell }_{t}}-\Delta \gamma )-s}{6cs-2s+{{\ell }_{t}}-\Delta \gamma } & \dfrac{2{{z}_{c}}}{6cs-2s+{{\ell }_{t}}-\Delta \gamma }  \\
%-\dfrac{\sqrt{3}s}{2s-{{\ell }_{t}}+\Delta \gamma } & \dfrac{2({{\ell }_{t}}-\Delta \gamma )-s}{6cs-2s+{{\ell }_{t}}-\Delta \gamma } & \dfrac{2{{z}_{c}}}{6cs-2s+{{\ell }_{t}}-\Delta \gamma }  \\
%\end{matrix} \right)
%\end{split}
\end{equation}
which is a specific version of the general Jacobian in (21).

We can obtain similar equations for all three scissor mechanisms. We denote the three Jacobians by ${{\mathbb{J}}_{S1}},{{\mathbb{J}}_{S2}},{{\mathbb{J}}_{S3}}$.

The apex local coordinates are functions of the end-effector position and orientation, ${{\mathbf{X}}_{\epsilon}}$ and ${{\Theta }_{\epsilon}}$, according to (17). For the first pair of scissors, its derivatives are given by
\begin{equation}
\Delta {{\mathbf{X}}_{C1}}=\Delta {{\mathbf{X}}_{\epsilon}}+{{\ell }_{t}}\frac{d\,\mathbf{\hat{n}}}{d{{\Theta }_{\epsilon}}}\Delta {{\Theta }_{\epsilon}}
\end{equation}
Substituting (27) in (25) yields
\begin{equation}
\left( \begin{matrix}
\Delta {{s}_{A1}}  \\
\Delta {{s}_{B1}}  \\
\end{matrix} \right)={{\mathbb{J}}_{S1}}\Delta {{\mathbf{X}}_{\epsilon}}+{{\ell }_{t}}{{\mathbb{J}}_{S1}}\frac{d\,\mathbf{\hat{n}}}{d{{\Theta }_{\epsilon}}}\Delta {{\Theta }_{\epsilon}}
\end{equation}
Similarly for the other two pairs of scissor mechanisms,
\begin{equation}
\begin{split}
& \left( \begin{matrix}
\Delta {{s}_{A2}}  \\
\Delta {{s}_{B2}}  \\
\end{matrix} \right)={{\mathbb{J}}_{S2}}\Delta {{\mathbf{X}}_{\epsilon}}-{{\ell }_{t}}{{\mathbb{J}}_{S2}}\left( \frac{\sqrt{3}}{2}\frac{d\,\mathbf{\hat{t}}}{d{{\Theta }_{\epsilon}}}+\frac{1}{2}\frac{d\,\mathbf{\hat{n}}}{d{{\Theta }_{\epsilon}}} \right)\Delta {{\Theta }_{\epsilon}} \\ 
& \left( \begin{matrix}
\Delta {{s}_{A3}}  \\
\Delta {{s}_{B3}}  \\
\end{matrix} \right)={{\mathbb{J}}_{S3}}\Delta {{\mathbf{X}}_{\epsilon}}+{{\ell }_{t}}{{\mathbb{J}}_{S3}}\left( \frac{\sqrt{3}}{2}\frac{d\,\mathbf{\hat{t}}}{d{{\Theta }_{\epsilon}}}-\frac{1}{2}\frac{d\,\mathbf{\hat{n}}}{d{{\Theta }_{\epsilon}}} \right)\Delta {{\Theta }_{\epsilon}} \\ 
\end{split}
\end{equation}
From (1) the derivatives of the unit vectors at the centerline are given by
\begin{equation}
\frac{d\,\mathbf{\hat{t}}}{d{{\Theta }_{\epsilon}}}=\left( \begin{matrix}
0 & 0 & 0  \\
0 & 0 & 1  \\
0 & -1 & 0  \\
\end{matrix} \right),\quad \frac{d\,\mathbf{\hat{n}}}{d{{\Theta }_{\epsilon}}}=\left( \begin{matrix}
0 & 0 & -1  \\
0 & 0 & 0  \\
1 & 0 & 0  \\
\end{matrix} \right)
\end{equation}
Combining these yields,
\begin{equation}
\Delta \mathbf{q}={{\mathbb{J}}_{t}}\Delta {{\mathbf{X}}_{\epsilon}}+{{\mathbb{J}}_{r}}\Delta {{\Theta }_{\epsilon}}
\end{equation}
where
\begin{equation}{{\mathbb{J}}_{t}}=\left( \begin{matrix}
{{\mathbb{J}}_{S1}}  \\
{{\mathbb{J}}_{S2}}  \\
{{\mathbb{J}}_{S3}}  \\
\end{matrix} \right)\in {{\Re }^{6\times 3}}\end{equation}
is the Jacobian associated with the translational displacement of the end-effector, and
\begin{equation}{{\mathbb{J}}_{r}}={{\ell }_{t}}\begin{pmatrix}
{{\mathbb{J}}_{S1}}\dfrac{d\,\mathbf{\hat{n}}}{d{{\Theta }_{\epsilon}}}  \\
-{{\mathbb{J}}_{S2}}\left( \dfrac{\sqrt{3}}{2}\dfrac{d\,\mathbf{\hat{t}}}{d{{\Theta }_{\epsilon}}}+\dfrac{1}{2}\dfrac{d\,\mathbf{\hat{n}}}{d{{\Theta }_{\epsilon}}} \right)  \\
{{\mathbb{J}}_{S3}}\left( \dfrac{\sqrt{3}}{2}\dfrac{d\,\mathbf{\hat{t}}}{d{{\Theta }_{\epsilon}}}-\dfrac{1}{2}\dfrac{d\,\mathbf{\hat{n}}}{d{{\Theta }_{\epsilon}}} \right)  \\
\end{pmatrix}\in {{\Re }^{6\times 3}}\end{equation}
is the Jacobian associated with the rotational displacement of the end-effector. From these we can obtain the block matrices (22) and (23), and thus the full Inverse Jacobian Matrix about this specific center point.

\subsection{Evaluation of Spatial Gear Ratios}
The maximum of the translational gear ratio ${{\gamma }_{t}}(\mathbf{p})=\left| \Delta {{\mathbf{X}}_{e}} \right|/\left| \Delta \mathbf{q} \right|$ is given by the minimum non-zero singular value associated with the Singular-Value Decomposition of Jacobian ${{\mathbb{J}}_{t}}$, (and vice-versa for the maximum gear ratio):
\begin{equation}{{\mathbb{J}}_{t}}={{\mathbf{U}}_{t}}{{\Sigma }_{t}}\mathbf{V}_{t}^{T}\end{equation}

where ${{\Sigma }_{t}}\in {{\Re }^{6\times 3}}$ is a rectangular diagonal matrix consisting of the square root of the eigenvalues associated with the real symmetric matrix $\mathbb{J}_{t}^{T}{{\mathbb{J}}_{t}}\in {{\Re }^{3\times 3}}$:
\begin{equation}
\begin{split}
&{{\Sigma }_{t}}=\left( \begin{matrix}
1/{{\lambda }_{t1}} & 0 & 0  \\
0 & 1/{{\lambda }_{t2}} & 0  \\
0 & 0 & 1/{{\lambda }_{t3}}  \\
0 & 0 & 0  \\
0 & 0 & 0  \\
0 & 0 & 0  \\
\end{matrix} \right)\\
&(0<{{\lambda }_{t1}}\le {{\lambda }_{t2}}\le {{\lambda }_{t3}})\end{split}\end{equation}
and ${{\mathbf{U}}_{t}}\in {{\Re }^{6\times 6}}$ and ${{\mathbf{V}}_{t}}\in {{\Re }^{3\times 3}}$ are, respectively, unitary matrices consisting of the eigenvectors of the matrices ${{\mathbb{J}}_{t}}\mathbb{J}_{t}^{T}\in {{\Re }^{6\times 6}}$  and $\mathbb{J}_{t}^{T}{{\mathbb{J}}_{t}}\in {{\Re }^{3\times 3}}$. Note that the Jacobian matrices we have obtained are for the \textit{inverse} kinematics relating actuator displacements to the end-effector displacements, thus taking the reciprocal of the eigenvalue in (30). The rotational gear ratio ${{\gamma }_{r}}(\mathbf{p})=\left| \Delta {{\Theta }_{e}} \right|/\left| \Delta \mathbf{q} \right|$can be examined in a similar manner.
\begin{figure}[ht]
	\centering
	\includegraphics[scale=.5]{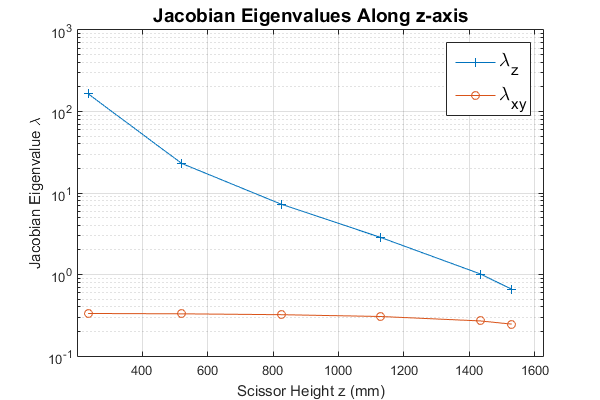}
	\caption{Jacobian at Various Heights along $z$-axis}
	\label{figurelabel}
\end{figure}

Fig. 8 
%and Fig. 9%
shows the translational gear ratio ${{\gamma }_{t}}(\mathbf{p})=\left| \Delta {{\mathbf{X}}_{e}} \right|/\left| \Delta \mathbf{q} \right|$ at diverse end-effector locations along the $z$-axis. Note that near the ground the vertical component of $\gamma_t$ grows large, and as the top platform nears the end of its upward travel $\gamma_t$ decreases, indicating a singular configuration where no more upward motion can be obtained. This can be interpreted intuitively by thinking of the three scissor mechanisms as the TSE top platform moves upward and how they elongate until no more motion can be achieved. 

%\begin{figure}[ht]
%	\centering
%	\includegraphics[scale=.58]{diagrams/EllipsoidsZTranslational}
%	\caption{Jacobian Ellipsoids for points along $z$-axis.}
%	\label{figurelabel}
%\end{figure}

\begin{figure}[ht]
  	\centering
  	\includegraphics[scale=.52]{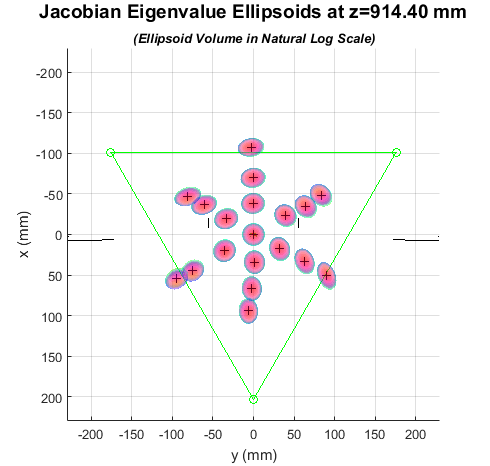}
  	\caption{Jacobian Ellipsoids for points at constant $z=914.40$ millimeters}
  	\label{figurelabel}
\end{figure}

Fig. 9 shows the translational gear ratio $\gamma_t$ for various points at a constant height in the center of the TSE's workspace. While virtual gear ratio values along the $x$ and $y$ axes do change as a function of the robot's configuration, they do not change as dramatically as the virtual gear ratio in the vertical direction does. This difference between the planar and vertical components of  $\gamma_t$ serves to further highlight the nature of the TSE, which was designed with vertical motion amplification in mind. This result serves to mathematically validate our design.

\section{Prototype}
A prototype Triple Scissor Extender (TSE) was built to test our kinematic model (see Fig. 10). The size of the robot was chosen such that it would have a maximum height of 1619.25 millimeters (63.75 inches) while being able to collapse to a height of 323.85mm (12.75 inches). The ratio between the lowest link length and the total length, $\ell_0/L$, is $0.0247$. This allows for a maximum height amplification of 5. 
%\begin{figure}[ht]
%	\centering
%	\includegraphics[scale=.5]{diagrams/Prototype}
%	\caption{Experimental Prototype}
%	\label{figurelabel}
%\end{figure} 
\begin{figure}[ht]
	\centering
	\includegraphics[scale=.325]{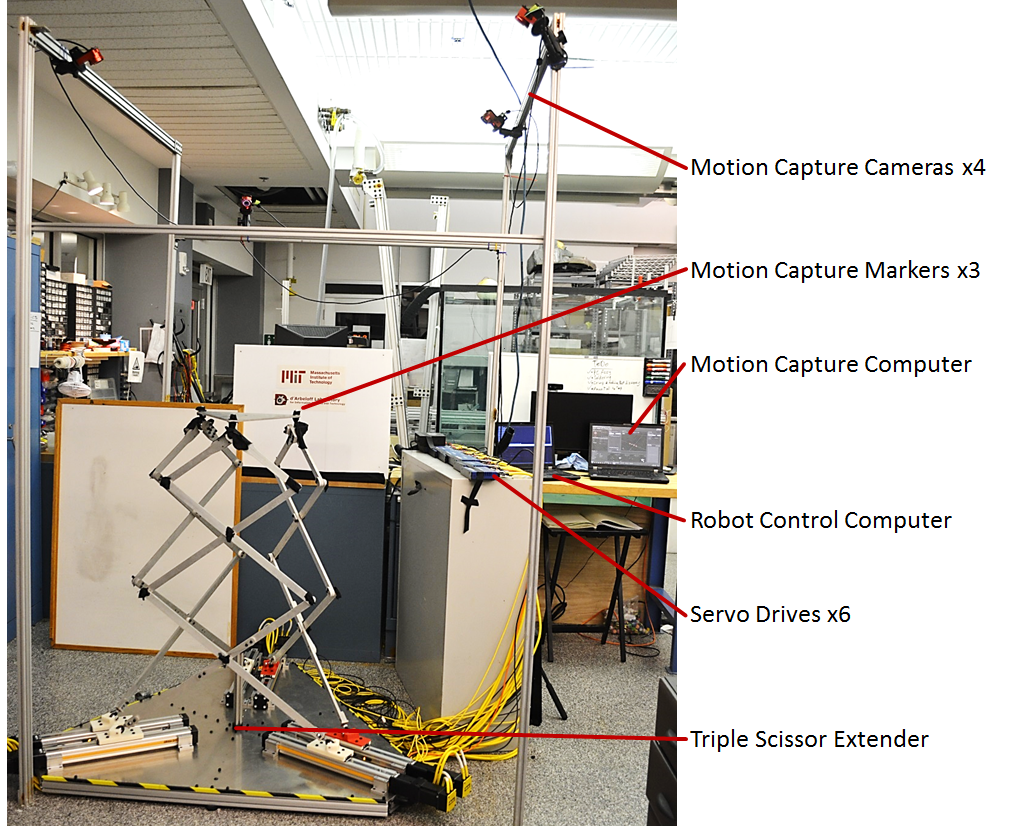}
	\caption{Experimental Setup}
	\label{figurelabel}
\end{figure} 

The ballscrew assemblies, servo motors, and servo drives were chosen by analyzing the worst-case loading while bearing a 11.34 kilogram (25 pound) gravitational load. The Parker Origa ballscrew servo assemblies are mounted on a single plate made from 6061 aluminum cut on an Omax waterjet. The plate has additional support from 8020 extrusion ribs and lining.

Each scissor mechanism is attached at the bottom points $A$ and $B$ as well as the top point $C$ by commercial off-the-shelf ball joint rod ends. The geometry was designed with hard stops such that the TSE would never pass through a singular configuration. The scissor links were all cut on a waterjet out of aluminum. Each joint uses twin flanged roller bearings that are preloaded for stiffness and a shoulder bolt as a pin. The top plate is also made of waterjet aluminum, and has mounting holes for the use of an end-effector or tool in the future. 

Each servo is driven and controlled by a Copley Accelnet panel. The six servo drives use a Kvaser Leaf CANopen USB Interface to communicate with a C++ control program running on a laptop running Ubuntu Linux 14.04LTS. This program leverages the C++ Motion Libraries (CML) provided by Copley for servo control and the Armadillo Linear Algebra Library \cite{RArmadillo} for computing the Inverse Jacobian $\mathbb{J}_{I}$ on-line. 

\section{Experimental Validation}
An experiment was conducted in order to evaluate the Inverse Jacobian's effectiveness for use in purely differential (stepping, or jogging) motion control of the TSE (See Fig. 10). By beginning the experiment at a known state of actuator displacements $q$ and top platform pose $p$, differential motion can be achieved by calculating the current Inverse Jacobian, multiplying it by a desired small change top platform pose, and commanding the actuators to move by that amount, as in (19). For simplicity, only the horizontal configuration (no rotation) was considered. 

The control program writes each new $p$ and $q$ to a file to create a dataset of desired states. The top platform's actual motion cam be measured using a motion capture system to create a dataset of actual states. An OptiTrack motion capture system consisting of four Flex3 cameras and a supporting structure was built around the TSE. OptiTrack Motive software was used to track markers that were placed at the three apexes of the TSE's top platform. These data were then exported and compared to the desired dataset to verify the Inverse Jacobian differential control scheme. 

\begin{figure}[ht]
	\centering
	\includegraphics[scale=.6]{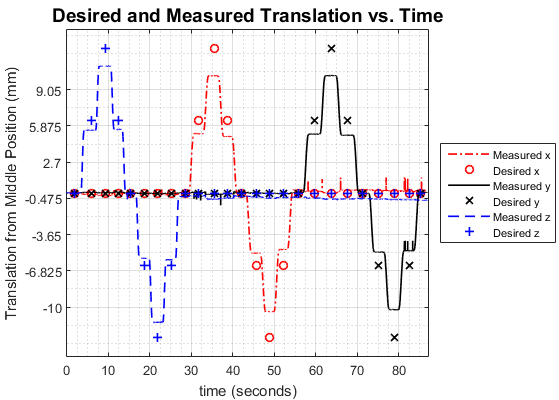}
	\caption{Top Platform Desired and Measured Translation vs Time}
	\label{figurelabel}
\end{figure} 

A set of discrete translations in increments of 6.35 millimeters (0.25 inches) was performed about a central pose at a height of 1173.16 millimeters (46.1875 inches), as shown in Fig. 11. 

\section{Discussion of Results}
Fig. 11 shows that the top platform consistently moves the same amount in each axis, but does not move the desired amount. This reveals a systematic error, either in the model or in the control architecture. The model does not take into account the passive structural mechanics such as the mass and stiffness of the individual components of the TSE and how gravitational loading may alter the modeled versus the actual position. 

Another possible source of error could be that the experimental jog amount of 6.35 millimeters is too large to allow for first-order control about the point. As the Jacobian is effectively a linearization about some operating point, this first-order approximation may not properly capture the nonlinear kinematics of the TSE at the scale of the jog amount. To resolve this, the jog amount would need to be decreased so that a new Inverse Jacobian calculated more often between motions, smoothing out the kinematic nonlinearities. 

Regardless of the cause of error, endpoint feedback using a motion capture system or some other external sensing could be used to obtain full closed-loop control of the TSE and eliminate any remaining error. The TSE using the Inverse Jacobian with either a passive structural mechanics model, fully continuous Jacobian calculation and motion integration, and/or closed-loop endpoint feedback control would be able to reach any pose within its workspace.

\addtolength{\textheight}{-12cm}   % This command serves to balance the column lengths
                                  % on the last page of the document manually. It shortens
                                  % the textheight of the last page by a suitable amount.
                                  % This command does not take effect until the next page
                                  % so it should come on the page before the last. Make
                                  % sure that you do not shorten the textheight too much.

%%%%%%%%%%%%%%%%%%%%%%%%%%%%%%%%%%%%%%%%%%%%%%%%%%%%%%%%%%%%%%%%%%%%%%%%%%%%%%%%

%%%%%%%%%%%%%%%%%%%%%%%%%%%%%%%%%%%%%%%%%%%%%%%%%%%%%%%%%%%%%%%%%%%%%%%%%%%%%%%%

%%%%%%%%%%%%%%%%%%%%%%%%%%%%%%%%%%%%%%%%%%%%%%%%%%%%%%%%%%%%%%%%%%%%%%%%%%%%%%%%

\end{document}